# Improving Urban Flood Prediction using LSTM-DeepLabv3+ and Bayesian Optimization with Spatiotemporal feature fusion

Zuxiang Situ[†, 1], Qi Wang[†, 1], Shuai Teng[1], Wanen Feng[1], Gongfa Chen[1],

Qianqian Zhou[*, 1], Guangtao Fu[*, 2]

[1] School of Civil and Transportation Engineering, Guangdong University of Technology,

No.100 Waihuan Xi Road, Guangzhou 510006, China;

[2] Centre for Water Systems, University of Exeter, Exeter EX4 4QF, United Kingdom

[†] These authors contributed equally to this work.

[*] Corresponding authors.

Email addresses: qiaz@foxmail.com (Q. Zhou), g.fu@exeter.ac.uk (G. Fu)

## Abstract

Deep learning models have become increasingly popular for flood prediction due to their superior accuracy and efficiency compared to traditional methods. However, current machine learning methods often rely on separate spatial or temporal feature analysis and have limitations on the types, number, and dimensions of input data. This study presented a CNN-RNN hybrid feature fusion modelling approach for urban flood prediction, which integrated the strengths of CNNs in processing spatial features and RNNs in analyzing different dimensions of time sequences. This approach allowed for both static and dynamic flood predictions. Bayesian optimization was applied to identify the seven most influential flood-driven factors and determine the best combination strategy. By combining four CNNs (FCN, UNet, SegNet, DeepLabv3+) and three RNNs (LSTM, BiLSTM, GRU), the optimal hybrid model was identified as LSTM-DeepLabv3+. This model achieved the highest prediction accuracy (MAE, RMSE, NSE, and KGE were 0.007, 0.025, 0.973 and 0.755, respectively) under various rainfall input conditions. Additionally, the processing speed was significantly improved, with an inference time of 1.158s (approximately 1/125 of the traditional computation time) compared to the physically-based models.

# 1 Introduction

Traditional flood prediction has been mainly based on simulations from process-based hydrodynamic models (Vozinaki et al. 2017, Apel et al. 2009, Jamali et al. 2018, Hosseiny 2021). Although the methods have been improved over the years, they still require extensive domain knowledge on complex construction procedures and a large variety of data that are difficult to collect (Noor et al. 2022, Han and Morrison 2022, Pham et al. 2021). Especially the surface inundation calculations need to solve a series of partial differential equations with conservation of mass and momentum and thus are computation intensive (Hosseiny 2021, Hofmann and Schüttrumpf 2021, Zheng et al. 2019). As a powerful artificial intelligence (AI) technology, Deep learning (DL) has gained growing popularity for hydrological modeling and flood predictions (Fu et al. 2022, Shen 2018). DL can extract complex features and capture nonlinear relationships hidden in large-scale data, with high accuracy and much lower computing cost (Han and Morrison 2022, LeCun et al. 2015, Rawat and Wang 2017).

Convolutional Neural Networks (CNNs) are one of the most successfully applied DL to learn spatial features (Hosseiny 2021, Hofmann and Schüttrumpf 2021, Zhang et al. 2022). CNNs were initially developed for computer vision tasks, and they can extract high-level features from spatial image data, such as satellite images, topographic features, photographs, and maps. This feature is significant as a majority of flood-driven factors are spatially distributed. CNNs can minimize the requirements of computing parameters by adopting partially connected layers and weight sharing strategy (Hofmann and Schüttrumpf 2021, LeCun et al. 2015). A few scholars employed CNNs in urban pluvial flood predictions, focusing on forecasting inundated regions and maximum water depths with inputs of unified rainfall and spatial features (e.g., DEMs and land use) (Fu et al. 2022, Guo et al. 2021, Löwe et al. 2021). However, CNNs require inputs of equal/fixed data dimensions and thus have been limited in analyzing dynamic sequence/length data (LeCun et al. 2015, Liu et al. 2020, Zhang et al. 2020).

Another research stream has focused on utilizing recurrent neural networks (RNNs) and their evolved versions to learn the temporal features from dynamic time series data (Zhang et al. 2022, Xu et al. 2021, Jiang et al. 2022). Different from CNNs, the RNNs adopt a unique feedback mechanism to convey information back to the network. When processing new data, all previously processed historical data are considered. Moreover, RNNs allow inputs with different dimensions or unequal lengths, which are more suitable for processing time-series data. The long short-term memory (LSTM) network has been mostly adopted thanks to its recurrent structure and unique gating mechanism. The method has shown great promise in hydrological predictions, such as flood warning and water level forecasting on river scale (Noor et al. 2022, Luppichini et al. 2022, Xu et al. 2020), rainfall-runoff predictions (Han and Morrison 2022, Zhu et al. 2020), flood susceptibility mapping (Fang et al. 2021, Wu et al. 2020), and identification of flood driven forces and mechanism (Jiang et al. 2022). However, the applications of LSTM to urban flood predictions, especially to forecasting dynamic changes of inundated space and depth, have been rare (Zhang et al. 2023). Despite RNNs' advantages in data dimensions, their prediction speeds are often slower than that of CNNs, especially when processing multiple types of large-scale spatial image features (Li et al. 2019, Hu et al. 2021).

Data fusion and hybrid models are recently proposed to optimize network capability in capturing both temporal and spatial features to overcome the limitations (Pham et al. 2021, Zheng et al. 2019, Zhang et al. 2022, Hong et al. 2018). Such methods can eliminate the weakness of a single data type or method by strategically combing two or multiple ones to achieve better model performance. In the field of urban flood prediction, there have been very limited studies exploring DL-based hybrid methods for water depth and dynamic predictions. The only two relevant studies both adopted the CNN-FNN (Feedforward Neural Network) model, in which the temporal characteristics (e.g., rainfall) were preprocessed and transformed by FNN in a simplified manner. Guo et al. (2021) employed a simple FNN to process rainfall hyetographs, which was further connected to a convolutional layer in the CNN model (i.e., FCN). Similarly, Löwe et al. (2021) preprocessed and converted rainfall sequences into a set of representative features before feeding them into an FNN model. The results were further integrated into an improved UNet network with skip connections. However, there has been no study exploring a CNN-RNN hybrid model, in which the RNN can be used to better identify the temporal features in time series data, discover rainfall patterns, and input key information into CNN to improve spatial feature understanding.

In the field of hydrological predictions in large river scales, CNN-RNN models have been increasingly adopted, nevertheless, their feature fusion has been implemented in two different ways: 1) the use of a ConvLSTM module, by combining convolution operations with LSTM to extract spatio-temporal information. Although this approach has improved flow prediction accuracy, it is still restricted in long-term prediction due to its complex calculation, large number of parameters and limited length of the input sequence (Chen et al. 2022). 2) establishing a CNN-RNN hybrid model by linking the RNN's output to the last layer of CNN. This is also the most commonly adopted approach for flow and runoff predictions (Chen et al. 2023, Li et al. 2022a, Li et al. 2022b, Xu et al. 2022). The model construction is more straightforward and less computationally intensive; however, it may lose some complex feature representations. How to achieve a deeper fusion of CNN and RNN that can allow for a richer feature representation and improved network's ability to introduce sequence information into deeper convolutional features is still a problem to be explored.

There is a critical gap in understanding the rationale of selecting driving factors for DL-based flood prediction models. Choosing appropriate driving factors is essential in improving network prediction accuracy, alleviating data collection burden and increasing processing speed (Pham et al. 2021, Zhang et al. 2022). However, in the past, input variables selection was mainly based on literature review, expert judgments, and correlation analyses (Guo et al. 2021, Löwe et al. 2021, Luppichini et al. 2022, Saha et al. 2022, Satarzadeh et al. 2022). The problem of those approaches is that the selection only considered the relationship between parameters, neglected the evaluation of their impacts on the model performance. Whether and to what extent the selected factors had actual impacts on DL models has been rarely explored. Optimization algorithms are thus necessary to address this issue and identify the optimal combination of input parameters.

Traditionally, genetic algorithms (GAs) have been used to optimize the structure of DL networks (Hong et al. 2018, Satarzadeh et al. 2022, Ronoud and Asadi 2019). The particle swarm optimization (PSO) has been increasingly explored to optimize the network architecture (such as the number of layers and the parameters of layers) of ML and DL methods in hydrological applications, such as flood susceptibility mapping (Pham et al. 2021) and flash flood segmentation (Tuyen et al. 2021). However, the GA and PSO need more initial sample points, and their calculation efficiency is low. Recently, Bayesian optimization (BO) has been developed as a novel algorithm that can automatically search for optimal hyperparameter combinations (Cui and Bai 2019). BO was used to obtain the parameter combination of the next iteration according to its prior probability and was found to have a higher optimization effectiveness and efficiency (Hebbal et al. 2022). However, there are currently no studies exploring the potential of BO in selecting the optimal combination of input driven factors.

In this study, we propose a novel CNN-LSTM hybrid model that enables spatiotemporal feature fusion for urban pluvial flood predictions. The contributions of this study are as follows:

(1) CNN and RNN are fully integrated for spatial and temporal features fusion for urban pluvial flood predictions. We constructed the CNN-RNN model differently by connecting the RNN's output to the deepest part of the CNN network. This allows for a more sophisticated capture and deeper fusion of spatiotemporal data. The method can make full use of CNNs' spatial processing capability to ensure processing speed and allow inputs of different data dimensions.

(2) Bayesian Optimization is employed in the hybrid model for selecting not only the number of input parameters, but also the specific types of parameters, based on the evaluation criteria of model prediction performance. The selected input factors are complementary in their spatial and temporal characteristics, which can effectively reduce the data collection burden and calculation demand in high-resolution flood forecasting tasks.

(3) In this study, 12 different types of CNN-RNN combined models were examined to determine the hybrid network that achieves the highest prediction accuracy with reasonable processing speed. This type of comparison is the first of its kind in urban flood prediction, as this study did not just use a specific combination strategy and/or its refinement to make predictions, but compared a variety of combinations. The optimal model was identified by comparing four CNNs and three RNNs that were combined and evaluated, which ensures its superiority and innovation.

## 2 Methodology

Fig. 1 presents an overview of the general methodology workflow, which includes three main steps: 1) a Bayesian optimization module to identify the optimal combination of input parameters/variables from a series of potential flood-driven factors; 2) a hybrid model construction module to develop a CNN-RNN combined network based on feature fusion approach; 3) a performance evaluation module, for comparison and selection of the best hybrid model based on a set of performance indicators. With the proposed methodology, the study can find the best input combination for the DL-based flood prediction model and identify the best combination approach for the hybrid network.

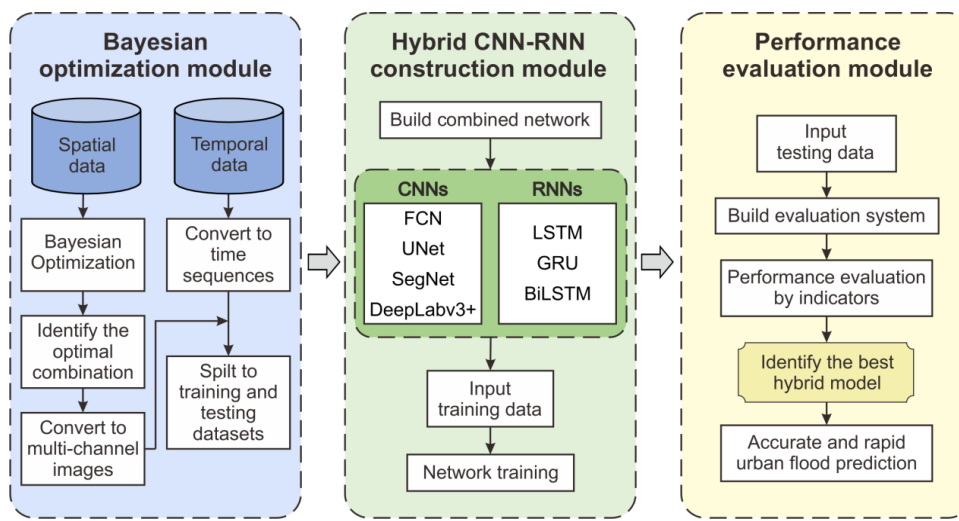

Fig. 1. An overview of methodology workflow

## 2.1 Selection of input spatial features

### 2.1.1 Potential flood-driven features

There are many urban flood-driven factors, such as rainfall characteristics, urban topography, land uses, and hydrological imperviousness. It is essential to identify and remove the ones with little or no predictive impacts on the DL-based flood model so that the model computation cost can be lowered and great efforts on data collection can be alleviated (Pham et al. 2021, Xu et al. 2021). In light of previous studies on catchment representation (Pham et al. 2021, Guo et al. 2021, Löwe et al. 2021, Satarzadeh et al. 2022), this study investigated 14 potential flood-driven features: DEM, ASP, CURV, DEM_L, SDEPTH, SLOPE, FLACC, TWI, IMP_C, IMP_S, IMP_SA, FLIMP, FLSLO, and PIPE. Detailed explanations of each of the variables are given in Table 1. Specifically, the first eight variables were generated based on the DEM in an ArcGIS environment. The imperviousness-related parameters were produced based on the regional land use and DEM. The PIPE layer describes the pipe network density and volume.

Table 1: Explanations of spatial features for DL-based flood prediction models

| **Abbr.** | **Variable** | **Explanations** |
|---|---|---|
| DEM | Digital Evaluation Model | Surface elevation, i.e., terrain model. |

| ASP | Aspect | The direction of water flow at each grid cell in the terrain data. It is the directional component of the gradient vector. |
|---|---|---|
| CURV | Curvature | The acceleration and deceleration of the flow, and the convergence and divergence of the flow, i.e., the concavity/convexity of the terrain. Cube root transformation is applied to reduce the extremely leptokurtic distribution of values. |
| DEM_L | DEM minus the local average of terrain elevation | DEM minus the focal mean of DEM within the 100 m radius. Pluvial urban flooding is linked to spatial scales <1 km (Löwe et al. 2021) and should be linked to local variations in elevation rather than elevation above sea level. |
| SDEPTH | Water depth | Water depth in terrain sinks is computed as the difference between the elevation of the outlet point of a sink and the terrain elevation and recorded as zero for cells located outside of sinks. |
| SLOPE | Slope | The magnitude of the gradient vector at each raster cell, representing the maximum rate of change in value from the center cell to its neighbors. It reflects the terrain's steepness and the overall movement of water. It was computed based on the focal mean of terrain elevation within the 100 m radius. |
| FLACC | Flow accumulation | The number of cells flowing into a given pixel describes the likelihood of a depression to be flooded. An upper cutoff at 1 ha is often defined as the threshold associated with natural streams. |

| | | Cube root transformation is applied to reduce the extremely leptokurtic distribution of values. |
|---|---|---|
| TWI | Topographic Wetness Index | Characterize the effect of regional topography on runoff flow direction and storage, and is a function of slope and upstream contributing area. Square root transformation is applied to reduce the extremely leptokurtic distribution of values. |
| IMP_C | Imperviousness of each raster cell | The impervious ratio of each pixel, and affects the amount of runoff generated. It is computed based on local building and road data that are assumed with an imperviousness of 100%. |
| IMP_S | Imperviousness of each subcatchment | The impervious ratio of each subcatchment and is computed based on land use distributions. |
| IMP_SA | IMP_S weighted by area | Imperviousness of each subcatchment is weighted by its area, and obtains a new imperviousness factor. |
| FLIMP | FLACC weighted by imperviousness in each pixel | Describing the total impervious area upstream from a given cell. It is computed by weighting the FLACC with IMP_C. An upper cutoff of 35 ha is set, with Cube root transformation applied. |
| FLSLO | FLACC weighted by the SLOPE in each cell | It is used to express average flow velocity on the path towards a cell that could quantify the ratio between infiltration and runoff. An upper cutoff at 10 ha is defined, and Cube root transformation is applied. |
| PIPE | Pipe network | The pipe network density and volume, thus reflecting the drainage capacity. |

**2.1.2 Bayesian optimization for selection and combination of input variables**

Due to the complex mechanism of urban flooding, there can be many input combinations due to the diverse types of potential inputs. Using all the potential parameters will reduce the network prediction speed, and some can even diminish the prediction accuracy. It is necessary to find the optimal input combination with an effective optimization method so that the DL model performance can be truly improved (Xu et al. 2021, Luppichini et al. 2022). This paper uses Bayesian Optimization (BO) to automatically search for the optimal combination of network input, thus improving the accuracy and efficiency of flood predictions. BO is a method that uses the Bayesian theorem to guide the search to find an objective function's minimum or maximum value and can be represented by a continuously updated probability model (see Supplementary Materials, Seq.1). The algorithm mimics how people learn by trial and error and then finding the best strategy based on those experiences the next time (see an overview of BO process in Supplementary Materials, SFig.1).

The BO parameter selection is to update the posterior distribution (Gaussian process) of a given optimization objective function by continuously adding sample points until the posterior distribution basically conforms to the actual distribution. The last parameter's information is considered to adjust the current parameter. Specifically, the selection process is in Eq. 1 and includes five steps (Fig. 2): (1) Define the range of the parameters, i.e., the 14 flood-driven factors listed in Table 1; (2) The BO algorithm infers the potential parameter combination; (3) The inferred combination is input into the DL models to predict floods; (4) Calculate the difference (i.e., error) between the predicted and actual values (i.e., ground truths); (5) The BO algorithm infers the parameter combination in the next iteration according to the error. Finally, after a series of iterative calculations, the BO can select a group of optimal parameter combinations (i.e., minimum error).

$$y^* = arg \min_{y \in \chi} f(y) \tag{1}$$

Where $f(y)$ is the objective function (prediction error), $y^*$ is the parameter combination, and $\chi$ is the potential range of the parameters.

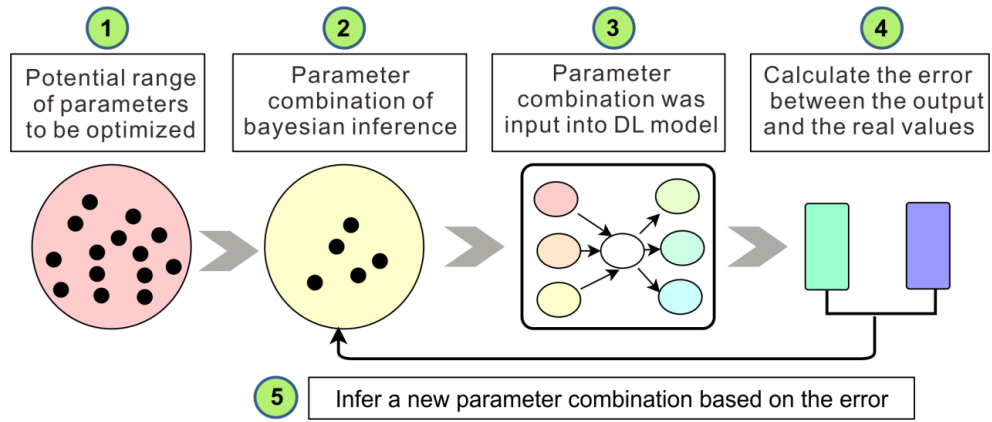

Fig. 2. Optimization process of flood-driven factors

## 2.2 Hybrid CNN-RNN for spatio-temporal feature fusion

### 2.2.1 Convolutional Neural Networks (CNNs) for spatial feature learning

CNN is a type of artificial neural network widely used in computer vision applications. The CNNs, from AlexNet (Krizhevsky et al. 2017) to Mask R-CNN (He et al. 2017), have proven effective in tasks such as image and vision capturing, learning, and analysis.

A typical CNN consists of a series of hidden layers, such as 1) the convolutional layer, which uses a convolutional kernel or filter to extract meaningful features from input or previous layer; 2) the pooling layer, which generalizes the existence of features and reduces data size by combining the output of neuron clusters from the previous layer into a single neuron; 3) batch normalization layer, normalizes the activations of the previous layer to a standard normal distribution by adjusting and scaling; 4) activation function layer, performs non-linear transformations to input features; and 5) fully connected layer, processes the features as a flattened matrix. With these layers and their functions, CNNs can significantly reduce the need for manual feature engineering (Sultana et al. 2020, Yu et al. 2019) and achieve efficient extraction capability (Simonyan and Zisserman 2014, He et al. 2016).

In this study, the CNN-based prediction process employed image semantic segmentation technology. Each pixel of the output images (i.e., flood inundation maps) is assigned a particular predicted feature or class, such as flood depth, based on input variables. A typical image segmentation network consists of two main components: an encoder and a decoder. Note that the detailed architecture may differ significantly for different networks, thus resulting in different predictive effects. To achieve a satisfying flood prediction, four types of mainstream segmentation networks were adopted for comparison, including FCN (Long et al. 2015), SegNet (Badrinarayanan et al. 2017), UNet (Ronneberger et al. 2015) and DeepLabv3+ (Chen et al. 2018). The details of each network are explained below.

*Fully Convolutional Networks (FCN)*

FCN adopts a CNN as the encoder module for downsampling, with deconvolutional layers in the decoder module for upsampling, see Fig 3a. The encoder is built with several local connection layers such as convolution, pooling, and upsampling layers, but without any dense layers. This allows FCN to process images with any data size. The encoder module extracts the data features by progressively reducing the image resolution, while the decoder module rescales the features to the original image size and recovers the detailed information. Note that the downsampling process will result in a loss of spatial information. A skip connection (highlighted by the dotted arrow) is used to connect the downsampled features to the upsampled layers at different resolutions and restore the spatial information in the downsampled layer. In this way, the skip connection can combine contextual information with spatial information, thus efficiently improving the model's accuracy.

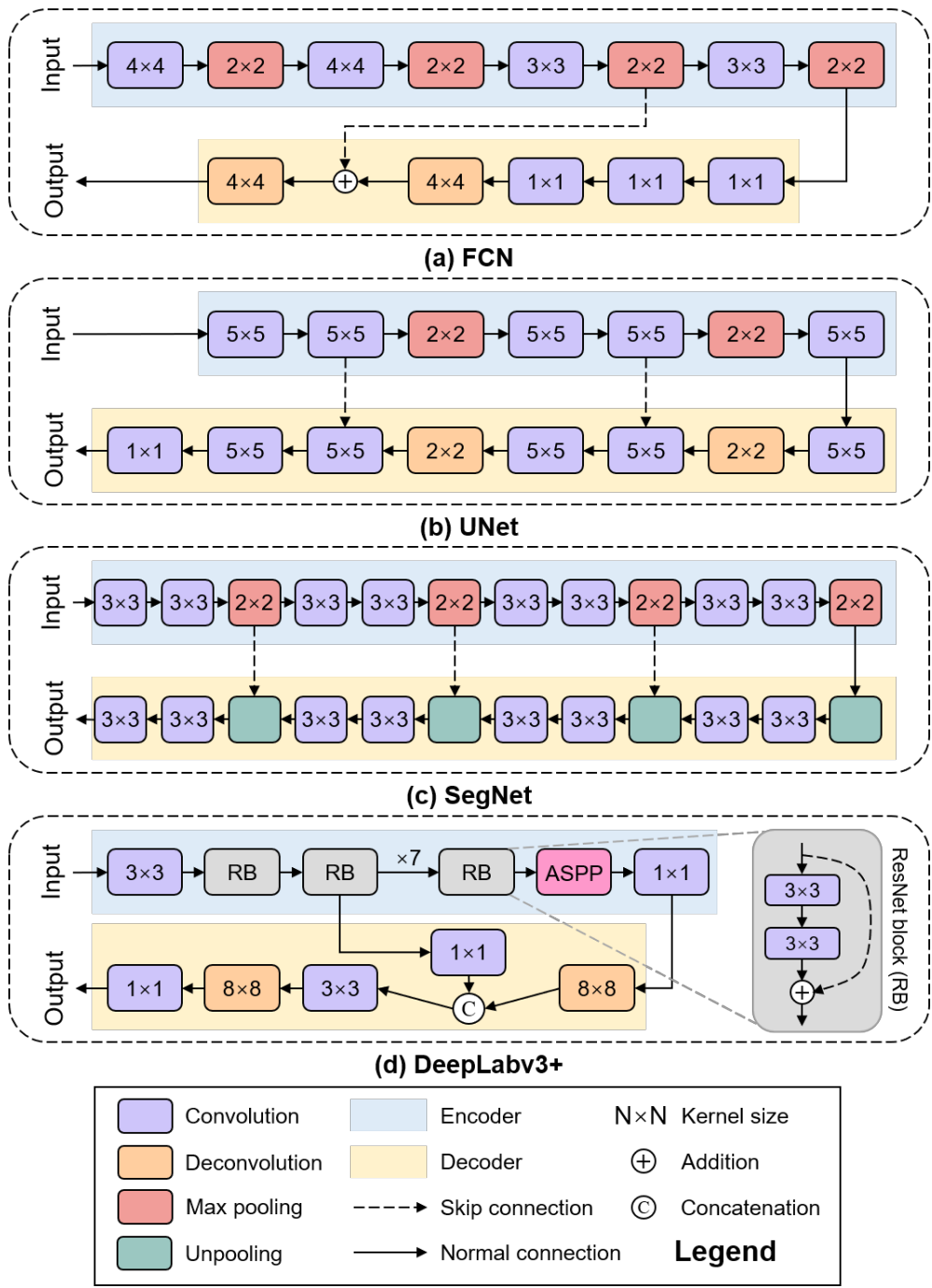

Fig. 3. FCN, UNet, SegNet, and DeepLabv3+ architecture

*UNet*

UNet is a variant of FCN and has been modified to improve the prediction performance with small datasets (Ronneberger et al. 2015, Zhao et al. 2019). UNet consists of a U-shaped encoder-decoder structure and is connected by two skip connections (Fig 3b). The encoder uses the convolutional layers to extract spatial features from input images. At the same time, the decoder performs upsampling to generate final prediction outputs. The output feature maps from the last 5×5 convolution operation in the encoder are passed to the decoder to perform upsampling using a 2×2 deconvolution operation. This upsampling process and two subsequent convolution operations are repeated twice, with the number of filters halved at each stage. The Rectified Linear Units (ReLU) are used as the activation function in all the convolution layers. The skip connections between the encoder and decoder enable the decoder to recover lost information during the pooling operations. The final prediction output is obtained through a 1×1 convolution operation.

*SegNet*

SegNet, the pixel-level fully convolutional neural network, has been developed specifically for flood prediction (Dong et al. 2021, Muhadi et al. 2021). The encoder module consists of eight convolutional layers, each corresponding to a decoder layer (Fig 3c). The convolutional layers apply batch normalization, ReLU activation function, and max-pooling operations. The decoder layers adopt convolutional, batch normalization, ReLU activation functions, and unpooling operations. SegNet only stores max pooling indexes and thus requires less memory. The decoder has 64 feature maps and thus requires a longer training process while providing more flexibility in obtaining higher training accuracy. The final output is a pixel-level classification, with each pixel associated with the predicted classes/values, such as flooded depths.

*DeepLabv3+*

DeepLabv3+ has been developed to deal with the problem of information loss in images due to reduced sizes of input feature maps during convolution and pooling processing. The main innovations are dilated convolution and Atrous Spatial Pyramidal Pooling (ASPP). The dilated convolution expands the receptive field of convolution kernels by inserting zeros between the filter values of each spatial dimension without increasing the computational resources. ASPP reduces the spatial information loss caused by pooling or convolution operations and expands the spatial context considered at each layer. As shown in Fig 3d, the network consists of three main components: a backbone network (i.e., ResNet18 constructed from the ResNet blocks (RB) with residual structure) for feature extraction, an ASPP module for per-pixel classification (including three parallel dilated convolution branches with dilation rates of 6, 12, and 18), and a 1×1 convolution after the ASPP module to produce an output. The encoder extracts basic information on the existence and location of the objects. The decoder then uses the extracted information to generate an output of the same size as the input.

### 2.2.2 Recurrent Neural Networks (RNNs) for temporal dynamic learning

In conventional feed-forward neural networks, the data stream transformation is passed in one direction through the hidden layer, and the current situation only influences the output. As a result, those networks have limited memory and are not suitable for modeling time-dependent data. To solve the limitation, Recurrent Neural Networks (RNN) have been developed to handle temporal data learning problems (Hochreiter and Schmidhuber 1997, Oksuz et al. 2019). When generating the output, RNNs consider past information's impact on it by using a gate unit that incorporates historical observations.

A typical RNN network consists of several hidden layers: 1) two recurrent layers used to capture temporal dependencies in the sequence data. At each time step, the recurrent units in the layers receive inputs of the previous hidden state and, in the meantime, output the current hidden state; 2) two dropout layers, as a commonly used regularization technique to drop neurons with a certain probability, thus allowing each part of the network to be trained effectively; 3) an activation layer, adopts the function of LeakyReLU that can solve the "neuron death" problem of ReLU when the input is negative; 4) a fully connected layer, with the same function described in the CNN chapter; 5) a reshape layer, used to project and reshape the output, thus rearranging the dimensions of the output to obtain a new vector representation. Meanwhile, there are three types of evolved RNN models that are particularly effective in modeling the time-dependent relationships in temporal data in a wide range of application domains (Mellit et al. 2021, Gao et al. 2020, Cho et al. 2022), including the Long Short-Term Memory (LSTM) (Hochreiter and Schmidhuber 1997), Bidirectional LSTM (BiLSTM) (Schuster and Paliwal 1997), and Gated Recurrent Unit (GRU) (Cho et al. 2014). Detailed explanations of each model are given below.

*Long Short-Term Memory (LSTM)*

LSTM is a complex gated storage unit designed to solve the gradient vanishing problem that hinders the efficiency of traditional RNNs. The problem occurs when the gradient of the optimizer becomes too small during training, which causes the connection weights to update slowly and leads to slow or stagnant training. As these weights control the flow of information and affect the network output, if the weights remain unchanged during training, the model may suffer from overfitting or underfitting, resulting in performance degradation. LSTM overcomes the problem by incorporating different gates to control the information flow through the network, see Fig. 4a. These gates (including the input gate, forget gate, and output gate) are implemented using weighted sums of logic functions, and their weights can be learned through backpropagation (more explanations in Supplementary Materials, SEq.2-7). Specifically, the input and forget gates control the unit states, while the output gate generates outputs and describes the memory to be used. This gives LSTM a more extended memory than traditional RNNs, making it well-suited for handling time series data and capturing long-term dependencies.

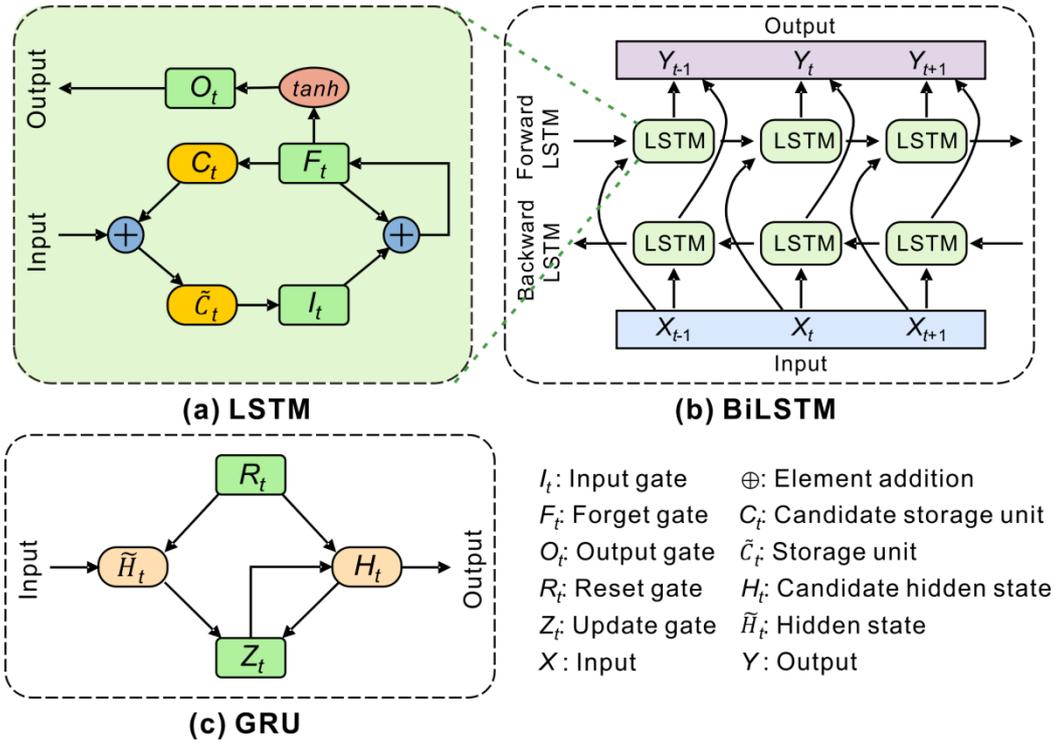

Fig. 4. LSTM, BiLSTM, and GRU architectures

*Bidirectional Long Short-Term Memory (BiLSTM)*

BiLSTM is an evolved version of the LSTM model. LSTM can capture the dependencies in sequential data by using hidden states and gates, but only based on the past context of the data up to the current state. The context of data that comes after the current state is also important in understanding the current state. To address this limitation, BiLSTM (see Fig. 4b) combines the bidirectional processing capabilities of RNNs with the ability of LSTM to capture dependencies in sequential data (Graves and Schmidhuber 2005). This allows BiLSTM to consider both past and future contexts when predicting the output of each time step, thereby improving the accuracy of the prediction. Specifically, BiLSTM uses two separate hidden states: one is to process the data forward, and the other is to process the data in the backward direction. The output from these two states is then combined to generate the final output at each time step (more explanations in Supplementary Materials, SEq.8-10).

*Gated Recurrent Unit (GRU)*

GRU is a variant of the LSTM model that aims to simplify the network design to improve the model performance. Unlike LSTM, GRU merges the input and forget gates into a single update gate (Fig. 4c), and therefore, there are only two gates (namely the update gate and reset gate) in GRU instead of three gates in LSTM. The change in the gates of GRU provides a new method for evaluating the hidden states. The update gate determines the number of previously stored memories to retain, while the reset gate works on combining the new input with the previous memories. The mathematical relationships between the different components of GRU are given in Supplementary Materials, SEq.11-14.

### 2.2.3 Spatial and temporal feature fusion strategy

This study proposed a new paradigm for feature fusion of spatial and temporal data through establishing a hybrid CNN-RNN model construction strategy (Fig. 5). The strategy differs from previous methods which have only been focused on extracting either temporal or spatial data. The methodology innovation of our approach is that various types of features with multiple dimensions can be input, processed, and learned simultaneously in the hybrid model. This enables a prediction of not only maximum water depth, but also the dynamic propagation in the time dimension, and with better model performance.

As shown in Fig. 5, there are two main streams of data: the spatial features are processed by the CNN network and the temporal ones by the RNN network and then concatenated in the CNN for post-processing analysis. The CNN recognizes spatial features (i.e., terrain, texture, edge, and pipeline) in images using convolutional and pooling layers. Specifically, all the spatial data are firstly stacked into multi-channel images through a pre-processing module and then input to the CNNs (i.e., FCN, UNet, SegNet, and DeepLabv3+) to extract key feature information (such as flow path and flood-prone area). On the other hand, the RNN network is used for recognizing temporal features in sequence data, such as changes in rainfall intensity and duration. The RNNs convert sequence features to a specified output size to connect with the features in the CNN in the depth dimension. Note that the concatenation positions of RNNs' output in the CNNs differ for the different CNN networks (see details in the marked red dotted box) but are the most feature-rich part (i.e., the narrowest part) of each network. The CNN model can understand the given temporal information with the temporal features combined.

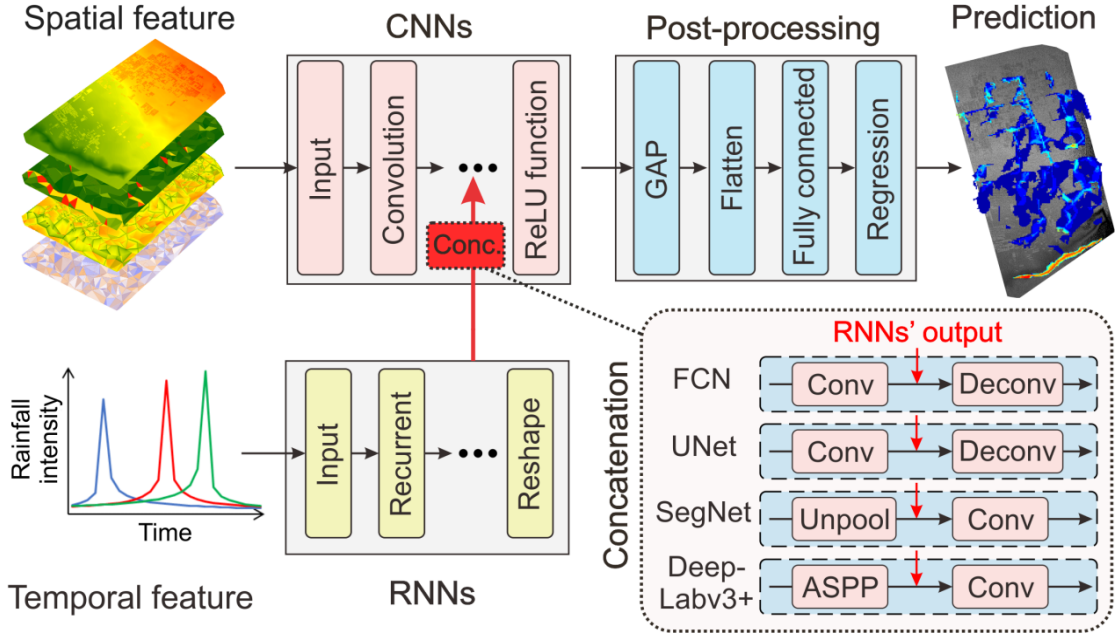

Fig. 5. Proposed hybrid CNN-RNN model architecture

After the output of RNNs is inserted into the CNNs for concatenation, the calculation will continue to complete the remaining layers of CNNs before proceeding to the next module, namely the post-processing. As shown in Fig. 5, it consists of four main layers, 1) the Global Average Pooling (GAP) layer, used to reduce the dimensionality of input features and improve computation speed; 2) the flatten layer, which aims to flatten the output from the GAP layer; 3) fully connected layer, transforms the output into a one-dimensional vector, with a length equal to the image resolution; and 4) the regression layer, used to output final prediction for urban flooding, i.e., the maximum flood depth or water depth at each time step for each pixel.

## 2.3 Performance indicators

To comprehensively evaluate hybrid models, prediction accuracy and computation time are taken into account. The prediction accuracy reflects the difference between ground truths (observed values) and predicted results. Four indicators are included, namely MAE (Mean Absolute Error), RMSE (Root Mean Square Error), NSE (Nash-Sutcliffe Efficiency) (McCuen et al. 2006), and KGE (Kling-Gupta Efficiency) (Gupta et al. 2009). The calculation time is reflected by two indicators: training time and inference time. Training time refers to the time required to complete one cycle of model training. Inference time is the time to predict an image, equivalent to the time to process an input raster. The units of the time measurement are all in seconds.

Specifically, MAE (Eq. 2) measures the degree of agreement between observed and predicted data in their actual units and evaluates all deviations from the observed values regardless of the event's magnitude. It is a non-negative indicator, and a result of zero indicates a perfect prediction. RMSE (Eq. 3) calculates the average error between the observed and predicted values by taking the square of the deviation. Similarly, a result of zero implies an ideal prediction. NSE (Eq. 4) has been widely used to evaluate performance of hydrological models. The range of NSE is between -inf and 1, with NSE=1 indicating the best fit. In hydrological modeling, NSE>0.5 is considered an acceptable level of performance (Knoben et al. 2019, Moriasi et al. 2007). KGE (Eq. 5) is calculated based on the decomposition of the Nash-Sutcliffe coefficient and mean squared error into three different terms representing bias, correlation, and the relative change between simulated and predicted values, respectively. KGE ranges from -inf to 1, with an ideal value of 1.

$$MAE = \frac{1}{n}\sum_{i=1}^{n}|Q_{obs} - Q_{sim}| \qquad (2)$$

$$RMSE = \sqrt{\frac{1}{n}\sum_{i=1}^{n}|Q_{obs} - Q_{sim}|^2} \qquad (3)$$

$$NSE = 1 - \left[\frac{\sum_{i=1}^{n}(Q_{obs} - Q_{sim})^2}{\sum_{i=1}^{n}(Q_{obs} - \overline{Q_{obs}})^2}\right] \qquad (4)$$

$$KGE = 1 - \sqrt{(r-1)^2 + (\alpha-1)^2 + (\beta-1)^2} \qquad (5)$$

Where $n$ is the number of samples, $Q_{obs}$ and $Q_{sim}$ represent the observed and predicted values, respectively. $\overline{Q_{obs}}$ represents the average of observed data. $r$ is the Pearson correlation coefficient, $\alpha$ is the ratio between the mean simulated and mean observed water depths, and $\beta$ denotes the bias.

## 3 Case study and model setting

### 3.1 Case study data

The case study is located in Hohhot, the capital of the Inner Mongolia Autonomous Region, northern China. The city is in a cold semi-arid climate zone with annual mean precipitation of approximately 396 mm (Zhou et al. 2018, Zhou et al. 2016). Land use can be categorized into mainly residential and commercial areas, institutes, green spaces, and other types. An illustration of the spatial driven variables in the area is shown in Fig. 6. The terrain changes from high to low in the north-south direction. The drainage system's capacity is insufficient in most areas, which cannot meet the once-a-year return period service level. It is noted that some spatial data exhibited a long-tailed and a right-skewed distribution (i.e., the data is unbalanced). However, the flood predictions did not change significantly due to these extreme values. Statistical transformations (i.e., square/cubic root transformation) were applied to bring the data closer to a normal distribution to prevent the modelling from overly influencing by the skewed data during training. Finally, all the adjusted spatial feature maps are stacked to create a multi-channel dataset with the same number of channels as variables.

For the temporal features, 90 rainfall events were used to simulate flood depths under different rainfall intensities with return periods ranging from 2 to 100 years. The rainfall hyetographs were generated following the national Technical Guidelines for the Establishment of Intensity-Duration-Frequency Curve and Design Rainstorm Profile (Zhou et al. 2018, MOHURD 2014, Berggren et al. 2014). The rainfall durations were set to 2, 4, and 6 hours, with a uniform 10-minute temporal resolution. Among the 90 rainfall events, 90% were randomly selected for model training and the rest 10% for testing. The predicted water depth maps are two-dimensional images in time series, with a spatial resolution of 231×125 pixels. For pixels with no data, i.e., no input data or flood depth, a value of zero was used for data padding.

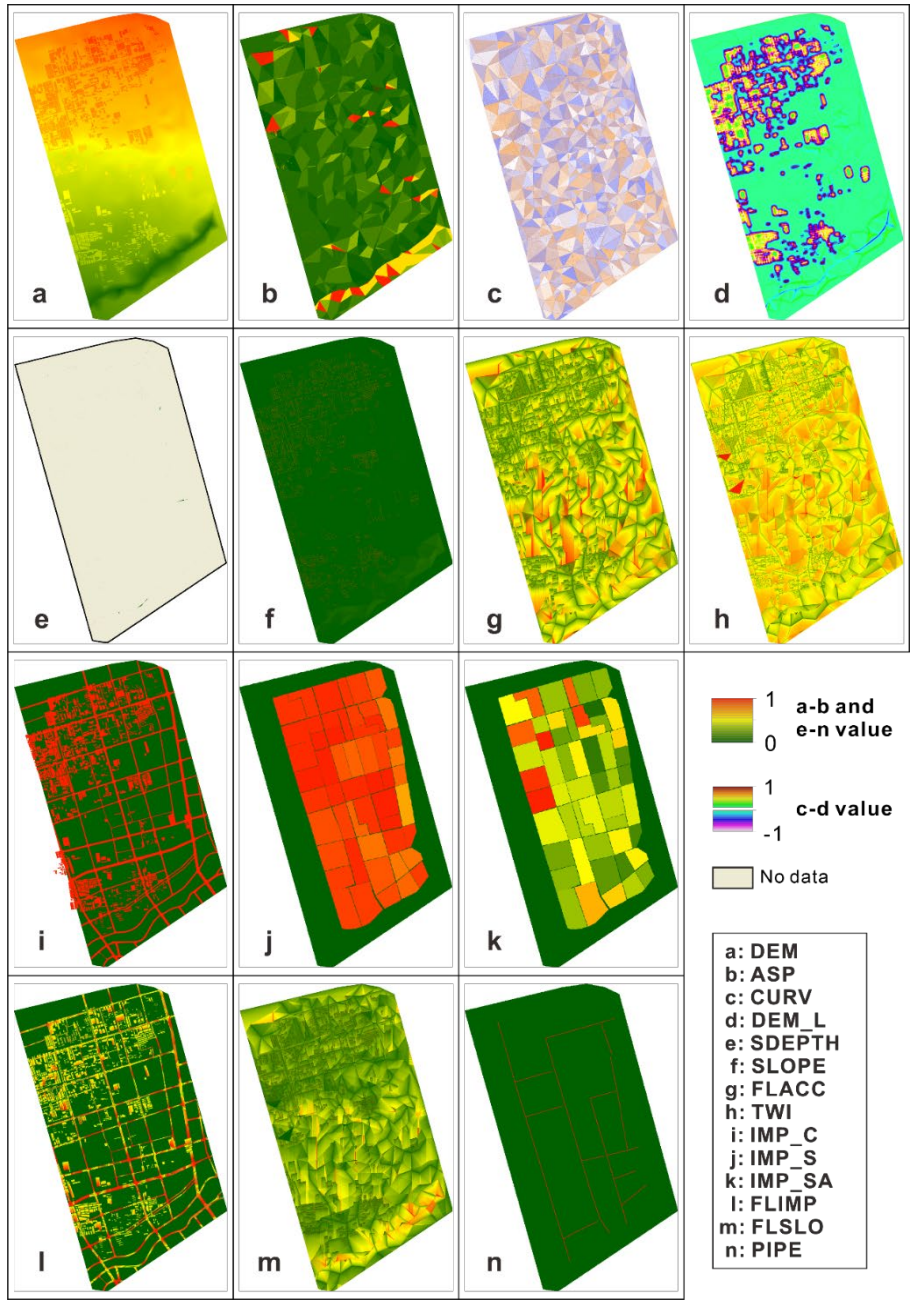

Fig. 6. Illustration of 14 input spatial-driven features

Meanwhile, data pre-processing was adopted to convert the different feature maps to match the size and resolution of the flood maps. All spatial and temporal data were also normalized by applying the linear min-max normalization method to map the feature values between 0 and 1 (excluding CURV and DEM_L, which were mapped between - 1 and 1). In doing so, the different features can have a consistent range of values, which helps the model better learn the features of data and speed up the training process.

## 3.2 1D&2D coupled hydrodynamic simulation

All the input flood maps (i.e., ground truths) were simulated by Mike Urban, a 1D&2D coupled hydrodynamic model that enables the simulations of underground pipe flow and overland inundation (MIKE by DHI 2016). The basic model inputs included the hydrological loads (e.g., rainfall time series), catchment characteristics (e.g., areas and imperviousness), and pipe network parameters (e.g., diameters, lengths, and elevations). The pipe flow was calculated using the Saint Venant equations, integrated with continuity and momentum equations. Once the pipe system was surcharged, the excess flow was conveyed to the surface for inundation calculations through coupled links established between the underground and surface systems in the MIKE 21 rectangular grid component. The model outputs include 1D pipeline results (e.g., pipe flow descriptions, surcharged manholes) and 2D overland results (e.g., maximum flood depths (maxH), flow paths, depths and velocities at different time steps). Among that, maxH has been mainly used to represent flood conditions (MIKE by DHI 2016, Kaspersen et al. 2017, Zhou et al. 2012). Samples of simulated flood maps can be found in Supplementary Materials, SFig.2.

## 3.3 Deep learning model setting and platform

The Bayesian optimization and the establishment of hybrid CNN-RNN models were implemented in MATLAB 2022b. The objective function of BO was defined as the minimum value of the RMSE. The number of optimization was 100 iterations. All the hybrid models were trained using the Adam optimizer for 100 epochs, which has been shown to achieve good stability and fast convergence for deep learning models (Luppichini et al. 2022, Chen et al. 2022). The momentum parameters, namely beta1, and beta2, were 0.9 and 0.999, respectively. The initial learning rate and mini-batch size were 0.01 and 8, respectively. To prevent gradient explosion, the gradient threshold was set to 1. The loss function of the training model was a mean squared error (MSE), and the pixels with no data were excluded from the evaluation of the loss function. The experimental environment was Windows 10, and the hardware configuration was Intel Xeon E5-2690 v4 CPU and NVIDIA Quadro P5000 GPU (16 GB memory).

## 4 Results and discussion

### 4.1 Key input variables identified by Bayesian optimization

Fig. 7 shows the impacts of the number of parameters involved in the combination on the model performance in the Bayesian optimization, thus identifying the optimal number for driven factor combinations. Specifically, subplots (a) and (b) illustrate the obtained MAE and RMSE as a result of different numbers of combined parameters, where the lower bars represent better results. In contrast, subplots (c) and (d) describe the achieved NSE and KGE, where a higher value is desired. The edge of the bar chart shows the mean value, and the error bar shows the standard deviation. Results show that MAE, NSE, and KGE had the best optimization effect when the number of parameters was set to 7. Only for RMSE, the optimal number of combinations was 5; nevertheless, the achieved performance was slightly better than the effect obtained with the number 7. This implies that providing more variables as input data for model training does not necessarily lead to better predictions. Sometimes, the input variables are not always positively correlated, and too much input information may mislead the learning and cause a decline in prediction accuracy and speed. On the other hand, when the number of input parameters was too small, the prediction results worsened due to the lack of sufficient data for learning. Therefore, it is vital to select the appropriate type and number of input parameters to ensure the model's performance.

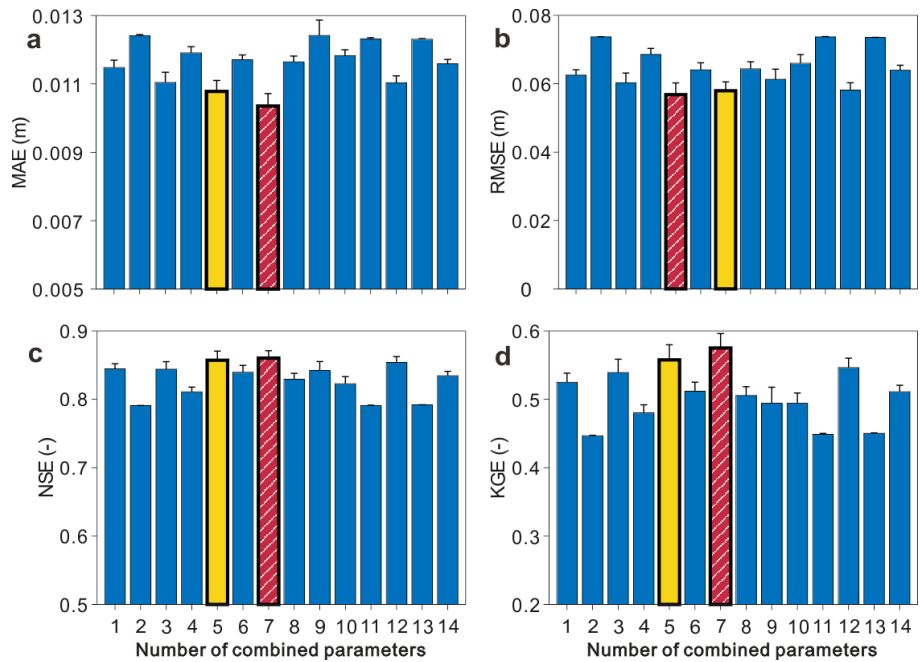

Fig. 7. Identification of the optimal number of parameters for input combination. The best and second-best options are highlighted in red (with diagonal lines) and bold

yellow boxes, respectively.

To further analyze which parameters had critical impacts on the predictive performance, we sorted the relevance of the 14 input variables based on the values of the four types of performance indicators in the order from best to worst in Fig. 8. The first seven parameters in each subplot (highlighted by the bold blue) indicate the most critical factors for the investigated indicator. It was shown that FLIMP and SDEPTH contributed the most to the best results in all four metrics, with ASP and IMP_C slightly behind in the rank and followed by PIPE and DEM. Note that DEM_L outperformed TWI in MAE but was inferior to TWI in the other three metrics. The seven identified key driving factors, namely DEM, ASP, SDEPTH, TWI, IMP_C, FLIMP, and PIPE, were consistent with the findings obtained when assessing the impacts of the potential driving factors on model performance (i.e., RMSE) over the 100 BO iterations (Supplementary Materials, SFig.3) and were used as input variables for the subsequent hybrid model construction.

Note that the four variables (DEM, ASP, SDEPTH, and TWI) generated based on DEM were beneficial to network training, but the other variables also caused harm to the network. This again addressed the need for parameter optimization to improve the model performance. Among impervious variables, IMP_C performed the best, indicating that describing the imperviousness in each cell was more effective in reflecting the impacts of land use on flood conditions. Moreover, it is also worth mentioning that the PIPE variable analyzed in this study was found to play a key role in flood prediction. This, however, was often neglected or less considered in previous studies.

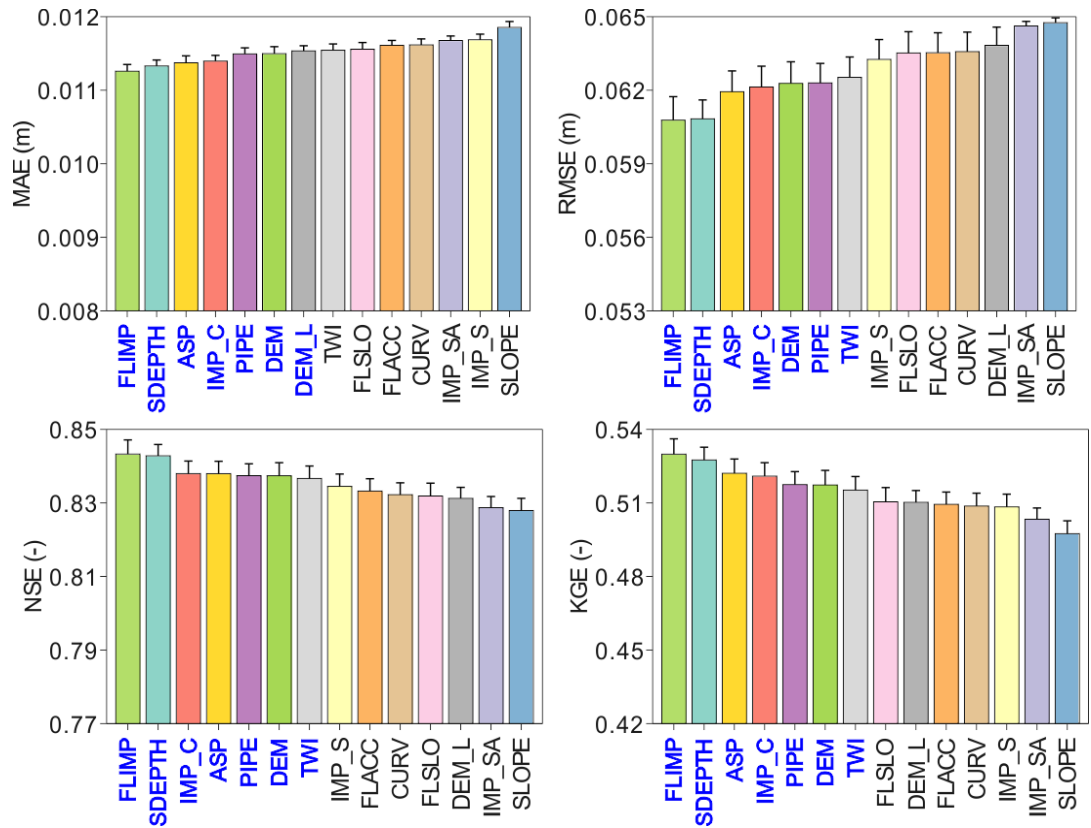

Fig. 8. The mean values and deviations of MAE, RMSE, NSE, and KGE obtained by the 14 potential drivers during the optimization process to reflect their contributions. The seven drivers that contribute the most are highlighted in bold blue.

## 4.2 Identification of the optimal hybrid model

Based on the seven selected input variables, different spatial and temporal prediction models were combined, and their prediction performances are summarized in Fig. 9. Since there were 4 CNNs and 3 RNNs, a total of 12 combinations were compared. The heat maps clearly show that the LSTM-DeepLabv3+ hybrid model had obvious advantages and was superior to the combinations of other models in terms of the four performance indicators. LSTM+DeepLabv3+ achieved the minimum error, which implies that the predicted flood depths in most of the cells were accurate. The high NSE and KGE hydrological index values also reflected the overall good performance of the model in flood predictions. Except for this optimal combination, LSTM+SegNet, or GRU+FCN/UNet also delivered satisfying results. For BiLSTM, most hybrid models did not work well though a combination with SegNet shows an acceptable performance.

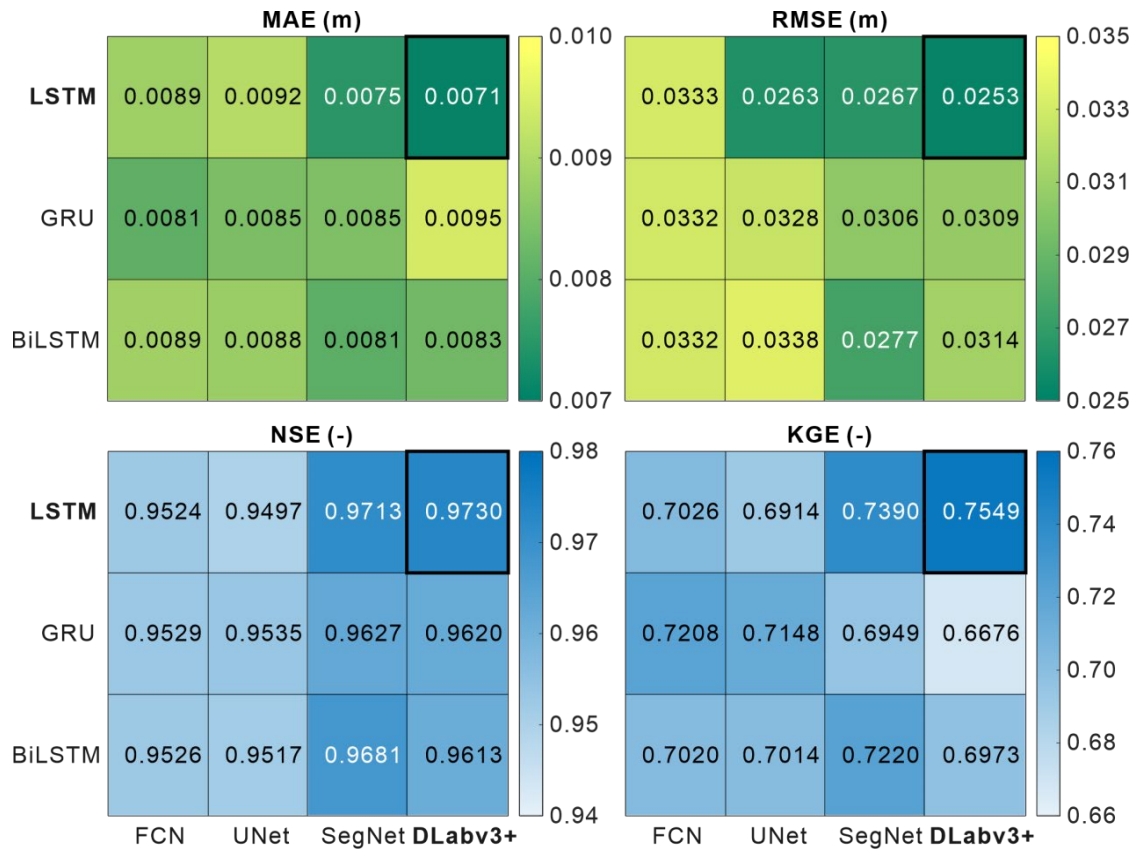

Fig. 9. MAE, RMSE, NSE, and KGE achieved by the different hybrid models. The black highlights the optimal combination of CNN-RNN for each performance index.

Fig. 10 shows the relationship among the training time, the number of parameters, and the required inference time of 12 hybrid models, with more details on the specific numerical results in Supplementary Materials, STal.1. Results showed that the size of the hybrid model was mainly influenced by the number of parameters of the CNNs (i.e., spatial feature prediction network), whereas the impacts of RNNs were much smaller. Meanwhile, the model size could significantly affect the training and inference time. For example, the hybrid models generated based on FCN had a large number of parameters, which required longer training and inference time. The SegNet was much more compact, with about a tenth of the size of the FCN, and thus the SegNet-based models generally processed faster. Nevertheless, the number of parameters did not entirely determine the processing time. For example, UNet had a unique network structure, and all the Unet-based hybrid models achieved the fastest inference speeds when combined with the different CNN networks. As for LSTM+DeepLabv3+ (i.e., with the highest prediction accuracy), the model processing speed was slightly slower, with an inference time of 1.158s, which was sufficient to meet the real-time prediction requirements.

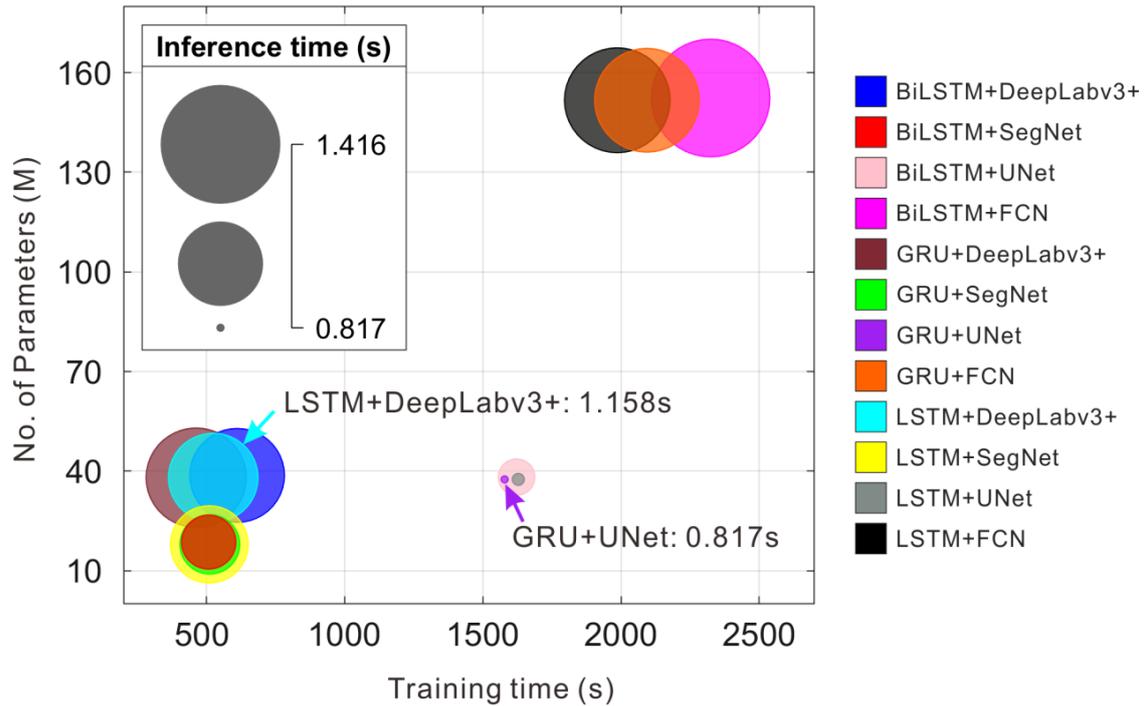

Fig. 10. The relationships among training time, number of parameters (in Millions), and inference time for the 12 hybrid models

Fig. 11 compares the flood predictions by all the hybrid models with a 70-year return period event. The specific RMSE and NSE values of each hybrid model are shown in Supplementary Materials, STal.2. The results were consistent with the previous findings that LSTM+DeepLabv3+ performed the best, with fewer areas of false predictions. This could be due to the fact that DeepLabv3+ adopted a fusion strategy of multiple features and thus have higher prediction accuracy. An increasing number of errors in the lower left corner were noted when LSTM was combined with the other three CNNs. The achieved model performance was relatively good when GRU was integrated with FCN and UNet. Nevertheless, many more missing predictions occurred when GRU was combined with DeepLabv3+, resulting in a much lower RMSE and NSE. The prediction results of BiLSTM-based hybrid models were generally not ideal. The reason could be that BiLSTM has a more complex structure with more parameters. When the dataset is small, the BiLSTM is easy to have an overfitting problem and leads to a decline in accuracy.

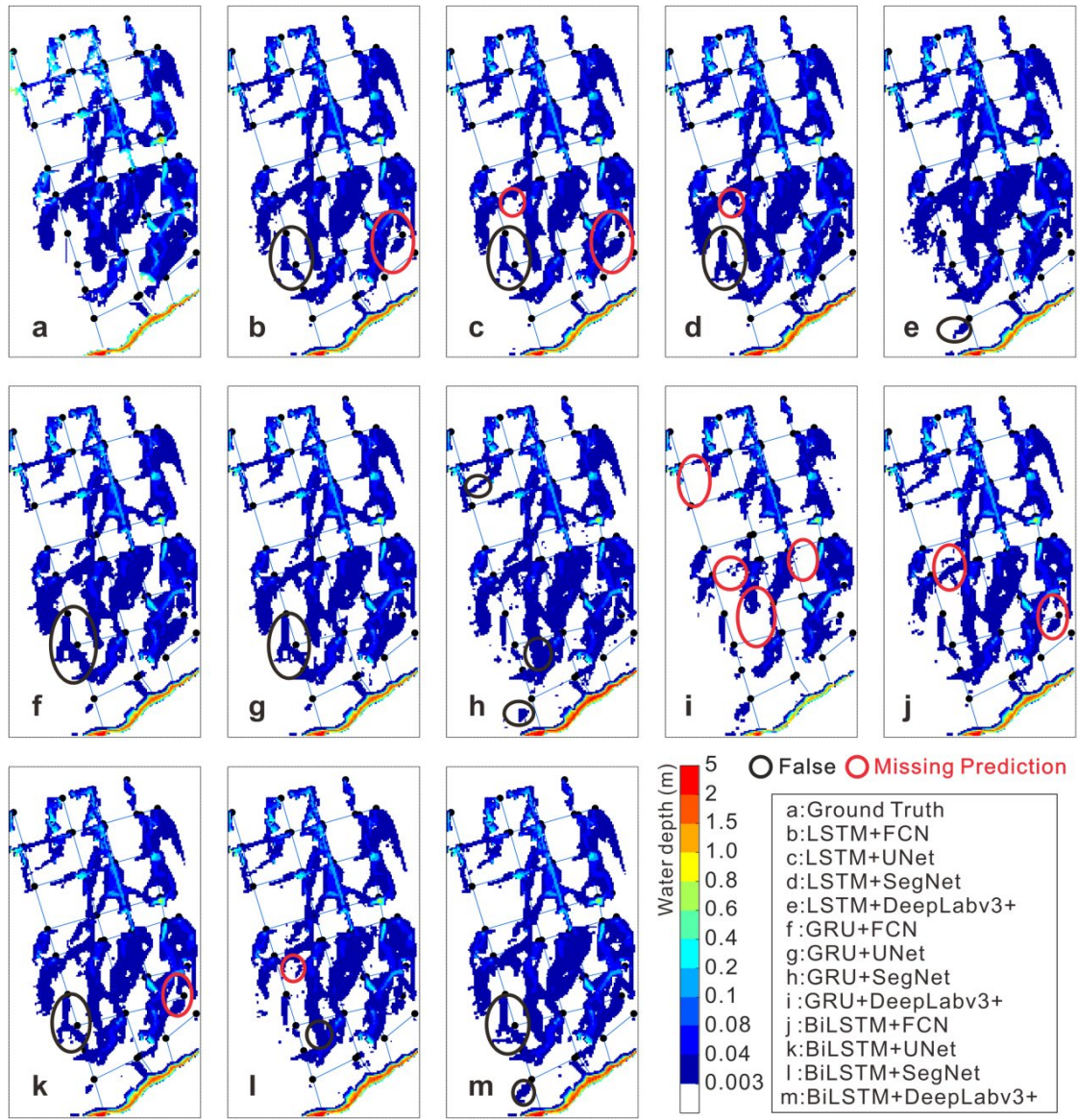

Fig. 11. Flood predictions by all hybrid models with a 70-yr event. The black and red circles highlighted the false and missing predictions, respectively.

## 4.3 Statistics of flood predictions by LSTM-DeepLabv3+

With the LSTM-DeepLabv3+ model, we further analyzed the correlation between the predicted water depths and the ground truths under different return periods (Fig. 12 a-i). With low return periods (i.e., below the 20-year event), the model tends to overestimate the flooded depths. For events between 20- and 60-year return periods, the correlation/agreement between model predictions and actual values reached the highest. Under extreme events, the model had conservative predictions and resulted in underestimated water depths. Fig. 12 j shows the changes in RMSE and NSE values with increasing return periods. The model performance could be further improved when the modelled rainfall was too small or too large (i.e., 2- or 100-year events). However, when taking into account all investigated return periods, on the whole, the model learned and performed well, achieving relatively low RMSEs and high NSEs under most events. More details on the correlation between water depths and performance indicators (i.e., RMSE and NSE) are presented in Supplementary Materials, SFig.4.

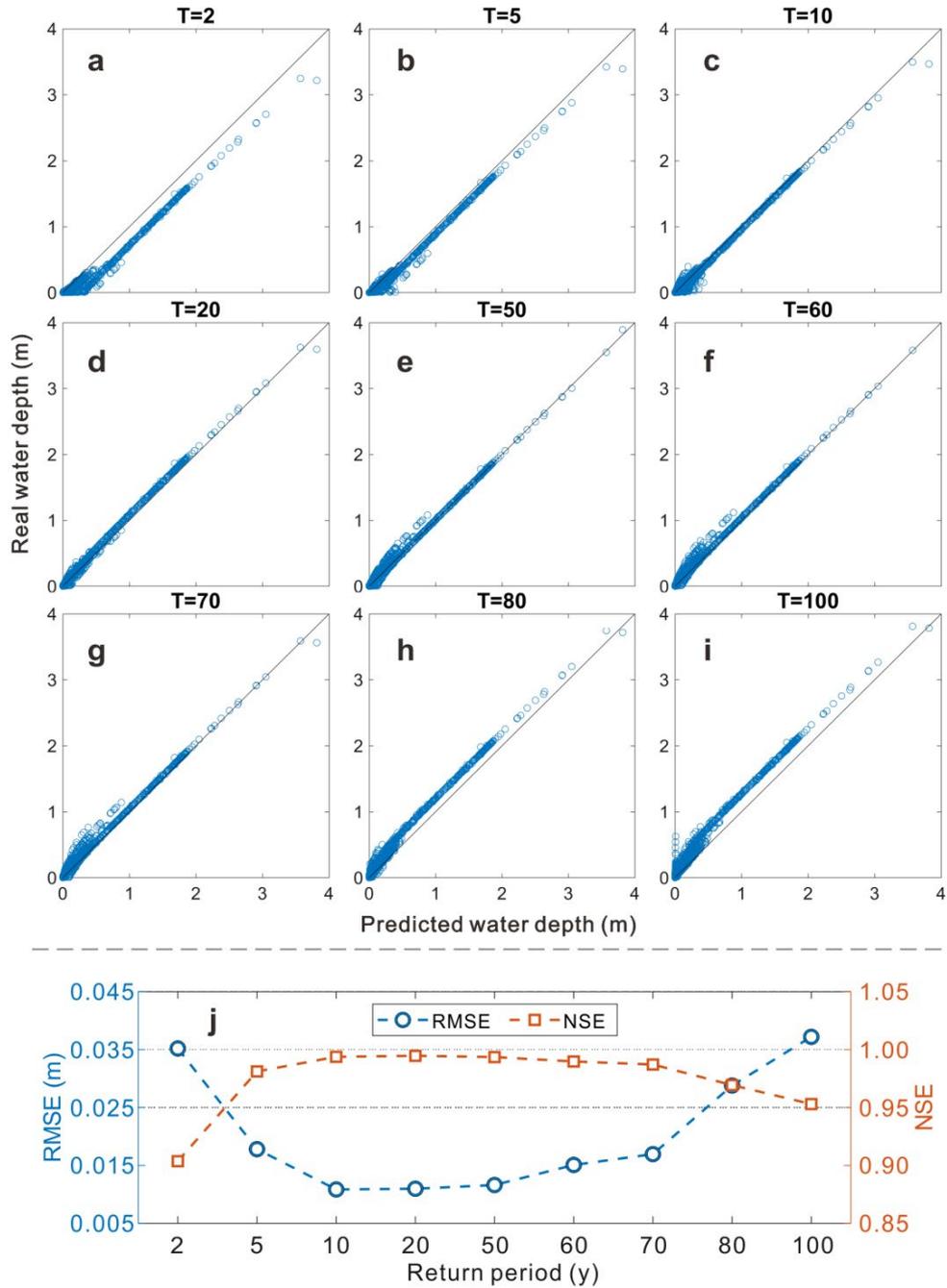

Fig. 12. (a-i) correlation between predicted depths and ground truths under 2-, 5-, 10-, 20-, 50-, 60-, 70-, 80-, 100- yr, and (j) RMSE and NSE as a function of return periods.

## 5 Conclusions

This study proposed a novel hybrid LSTM-DeepLabv3+ model for accurate and rapid urban flood prediction based on multi-dimensional spatial and temporal features. Our model fully leveraged the strengths of CNNs in processing spatial features and RNNs in processing different time dimensions of input factors to achieve accurate predictions under dynamic flood scenarios. To optimize the performance of the model, we employed Bayesian optimization that allows us to not only identify the most influential driving factors but also determine the optimal combination strategy for input parameters, including the type and number of inputs required to achieve the best results. Moreover, to have a comprehensive valuation, we tested and compared 12 different types of CNN-RNN hybrid models, which include four CNNs combined with three RNNs. The LSTM-DeepLabv3+ was identified as the best model with the highest prediction accuracy and reasonable processing speed. Key findings obtained from a study of Hohhot are summarised below.

Among the various types of spatial factors, it was found that seven variables had the most important impacts on flood predictions, including DEM, ASP, SDEPTH, FLIMP, TWI, IMP_C, and PIPE (see Table 1). the role of PIPE which was often neglected in previous research, was found to be important to improve predictive accuracy. The results showed that the adopted Bayesian optimization approach was essential to capture the combination strategy of these factors, as either too few or too many input variables could have an adverse impact on model performance. The Bayesian optimization was more reliable in selecting the types and number of parameters than the manual selections and managed to avoid the model being affected by many unreasonable choices.

Regarding the computational time requirements of hybrid models , it was found that the size and processing speed of the hybrid models were mainly influenced by the properties of CNNs. Therefore, the hybrid models generated based on FCN had slower processing speeds as the models were associated with a large number of parameters. Meanwhile, the network architecture could also affect the model's performance. All UNet-based hybrid models achieved low inference time as UNet had a unique network structure. Nevertheless, among all hybrid models, the combination of LSTM and DeepLabv3+ was selected because the model achieved the highest prediction accuracy and an inference time of 1.158s, which was sufficient to meet the real-time prediction requirements.

The flood predictions under different rainfall conditions showed the hybrid models achieve an overall good model performance and high robustness, with low RMSEs and high NSEs. At the same time, the results also showed that LSTM-DeepLabv3+ worked best in predicting flood conditions under return periods between 20 and 70 years. When input rainfall intensities were too small or extreme, the model tended to overestimate or underestimate the flood depths, resulting in slight deviations.

This work also has some limitations. The ground truths used in this study were obtained by the 1D-2D coupled hydrodynamic simulations rather than the actual field site measurements. This was because, in general, there is lack sufficient and good-quality data from field surveys. In future work, the hybrid models should be tested with monitored data. On the other hand, future work needs to consider integrating the proposed spatiotemporal hybrid model into urban flood warning and forecasting systems. Thanks to the high computational efficiency, LSTM-DeepLabv3+ could be used for real-time early warning to improve the evaluation of and response to urban flood events.

## Acknowledgments

This research was funded by the Youth Promotion Program of the Natural Science Foundation of Guangdong Province, China (Grant No: 2023A1515030126).

## References

Vozinaki, A.E.K., Morianou, G.G., Alexakis, D.D. and Tsanis, I.K. (2017) Comparing 1D and combined 1D/2D hydraulic simulations using high-resolution topographic data: a case study of the Koiliaris basin, Greece. Hydrological Sciences Journal-Journal des Sciences Hydrologiques 62(4), 642-656.

Apel, H., Aronica, G.T., Kreibich, H. and Thieken, A.H. (2009) Flood risk analyses-how detailed do we need to be? Natural Hazards 49(1), 79-98.

Jamali, B., Löwe, R., Bach, P.M., Urich, C., Arnbjerg-Nielsen, K. and Deletic, A. (2018) A rapid urban flood inundation and damage assessment model. Journal of Hydrology 564, 1085-1098.

Hosseiny, H. (2021) A deep learning model for predicting river flood depth and extent. Environmental Modelling & Software 145, 105186.

Noor, F., Haq, S., Rakib, M., Ahmed, T., Jamal, Z., Siam, Z.S., Hasan, R.T., Adnan, M.S.G., Dewan, A. and Rahman, R.M. (2022) Water Level Forecasting Using Spatiotemporal Attention-Based Long Short-Term Memory Network. Water 14(4), 612.

Han, H. and Morrison, R.R. (2022) Improved runoff forecasting performance through error predictions using a deep-learning approach. Journal of Hydrology 608, 127653.

Pham, B.T., Luu, C., Phong, T.V., Trinh, P.T., Shirzadi, A., Renoud, S., Asadi, S., Le, H.V., von Meding, J. and Clague, J.J. (2021) Can deep learning algorithms outperform benchmark machine learning algorithms in flood susceptibility modeling? Journal of Hydrology 592, 125615.

Hofmann, J. and Schüttrumpf, H. (2021) floodGAN: Using Deep Adversarial Learning to Predict Pluvial Flooding in Real Time. Water 13(16), 2255.

Zheng, Y.H., Song, H.H., Sun, L., Wu, Z.B. and Jeon, B. (2019) Spatiotemporal Fusion of Satellite Images via Very Deep Convolutional Networks. Remote Sensing 11(22).

Fu, G.T., Jin, Y.W., Sun, S.A., Yuan, Z.G. and Butler, D. (2022) The role of deep learning in urban water management: A critical review. Water Research 223.

Shen, C. (2018) A transdisciplinary review of deep learning research and its relevance for water resources scientists. Water Resources Research 54(11), 8558-8593.

LeCun, Y., Bengio, Y. and Hinton, G. (2015) Deep learning. Nature 521(7553), 436-444.

Rawat, W. and Wang, Z. (2017) Deep Convolutional Neural Networks for Image Classification: A Comprehensive Review. Neural Computation 29(9), 2352-2449.

Zhang, X.B., Jin, Q.Z., Yu, T.Z., Xiang, S.M., Kuang, Q.M., Prinet, V. and Pan, C.H. (2022) Multi-modal spatio-temporal meteorological forecasting with deep neural network. Isprs Journal of Photogrammetry and Remote Sensing 188, 380-393.

Guo, Z.F., Leitao, J.P., Simoes, N.E. and Moosavi, V. (2021) Data-driven flood emulation: Speeding up urban flood predictions by deep convolutional neural networks. Journal of Flood Risk Management 14(1).

Löwe, R., Böhm, J., Jensen, D.G., Leandro, J. and Rasmussen, S.H. (2021) U-FLOOD – Topographic deep learning for predicting urban pluvial flood water depth. Journal of Hydrology 603, 126898.

Liu, J., Yang, X., Lau, S., Wang, X., Luo, S., Lee, V.C.-S. and Ding, L. (2020) Automated pavement crack detection and segmentation based on two-step convolutional neural network. Computer-Aided Civil and Infrastructure Engineering 35(11), 1291-1305.

Zhang, C., Chang, C.-c. and Jamshidi, M. (2020) Concrete bridge surface damage detection using a single-stage detector. Computer-Aided Civil and Infrastructure Engineering 35(4), 389-409.

Xu, Y., Hu, C., Wu, Q., Li, Z., Jian, S. and Chen, Y. (2021) Application of temporal convolutional network for flood forecasting. Hydrology Research 52(6), 1455-1468.

Jiang, S., Zheng, Y., Wang, C. and Babovic, V. (2022) Uncovering Flooding Mechanisms Across the Contiguous United States Through Interpretive Deep Learning on Representative Catchments. Water Resources Research 58(1), e2021WR030185.

Luppichini, M., Barsanti, M., Giannecchini, R. and Bini, M. (2022) Deep learning models to predict flood events in fast-flowing watersheds. Science of The Total Environment 813, 151885.

Xu, W., Jiang, Y., Zhang, X., Li, Y., Zhang, R. and Fu, G. (2020) Using long short-term memory networks for river flow prediction. Hydrology Research 51(6), 1358-1376.

Zhu, S., Luo, X., Yuan, X. and Xu, Z. (2020) An improved long short-term memory network for streamflow forecasting in the upper Yangtze River. Stochastic Environmental Research and Risk Assessment 34(9), 1313-1329.

Fang, Z., Wang, Y., Peng, L. and Hong, H. (2021) Predicting flood susceptibility using LSTM neural networks. Journal of Hydrology 594, 125734.

Wu, Y., Ding, Y., Zhu, Y., Feng, J. and Wang, S. (2020) Complexity to Forecast Flood: Problem Definition and Spatiotemporal Attention LSTM Solution. Complexity 2020, 7670382.

Zhang, L., Qin, H., Mao, J., Cao, X. and Fu, G. (2023) High temporal resolution urban flood prediction using attention-based LSTM models. Journal of Hydrology, 129499.

Li, L.Y., Chen, Y., Xu, T.B., Shi, K.F., Huang, C., Liu, R., Lu, B.B. and Meng, L.K.

(2019) Enhanced Super-Resolution Mapping of Urban Floods Based on the Fusion of Support Vector Machine and General Regression Neural Network. Ieee Geoscience and Remote Sensing Letters 16(8), 1269-1273.

Hu, W., Wang, W., Ai, C., Wang, J., Wang, W., Meng, X., Liu, J., Tao, H. and Qiu, S. (2021) Machine vision-based surface crack analysis for transportation infrastructure. Automation in Construction 132, 103973.

Hong, H., Panahi, M., Shirzadi, A., Ma, T., Liu, J., Zhu, A.X., Chen, W., Kougias, I. and Kazakis, N. (2018) Flood susceptibility assessment in Hengfeng area coupling adaptive neuro-fuzzy inference system with genetic algorithm and differential evolution. Science of the Total Environment 621, 1124-1141.

Chen, C., Jiang, J., Liao, Z., Zhou, Y., Wang, H. and Pei, Q. (2022) A short-term flood prediction based on spatial deep learning network: A case study for Xi County, China. Journal of Hydrology 607, 127535.

Chen, J., Li, Y. and Zhang, S. (2023) Fast Prediction of Urban Flooding Water Depth Based on CNN−LSTM. Water 15(7), 1397.

Li, P., Zhang, J. and Krebs, P. (2022a) Prediction of Flow Based on a CNN-LSTM Combined Deep Learning Approach. Water 14(6), 993.

Li, X., Xu, W., Ren, M., Jiang, Y. and Fu, G. (2022b) Hybrid CNN-LSTM models for river flow prediction. Water Supply 22(5), 4902-4919.

Xu, C., Wang, Y., Fu, H. and Yang, J. (2022) Comprehensive Analysis for Long-Term Hydrological Simulation by Deep Learning Techniques and Remote Sensing. Frontiers in Earth Science 10.

Saha, S., Gayen, A. and Bayen, B. (2022) Deep learning algorithms to develop Flood susceptibility map in Data-Scarce and Ungauged River Basin in India. Stochastic Environmental Research and Risk Assessment 36(10), 3295-3310.

Satarzadeh, E., Sarraf, A., Hajikandi, H. and Sadeghian, M.S. (2022) Flood hazard mapping in western Iran: assessment of deep learning vis-à-vis machine learning models. Natural Hazards 111(2), 1355-1373.

Ronoud, S. and Asadi, S. (2019) An evolutionary deep belief network extreme learning-based for breast cancer diagnosis. Soft Computing 23(24), 13139-13159.

Tuyen, D.N., Tuan, T.M., Son, L.H., Ngan, T.T., Giang, N.L., Thong, P.H., Hieu, V.V., Gerogiannis, V.C., Tzimos, D. and Kanavos, A. (2021) A Novel Approach Combining Particle Swarm Optimization and Deep Learning for Flash Flood Detection from Satellite Images. Mathematics 9(22), 2846.

Cui, H. and Bai, J. (2019) A new hyperparameters optimization method for convolutional neural networks. Pattern Recognition Letters 125, 828-834.

Hebbal, A., Balesdent, M., Brevault, L., Melab, N. and Talbi, E.-G. (2022) Deep Gaussian process for multi-objective Bayesian optimization. Optimization and Engineering.

Krizhevsky, A., Sutskever, I. and Hinton, G.E. (2017) Imagenet classification with deep convolutional neural networks. Communications of the ACM 60(6), 84-90.

He, K., Gkioxari, G., Dollár, P. and Girshick, R. (2017) Mask r-cnn, pp. 2961-2969.

Sultana, J., Usha Rani, M. and Farquad, M. (2020) Emerging research in data engineering systems and computer communications, pp. 511-519, Springer.

Yu, Y., Wang, C., Gu, X. and Li, J. (2019) A novel deep learning-based method for damage identification of smart building structures. Structural Health Monitoring 18(1), 143-163.

Simonyan, K. and Zisserman, A. (2014) Very deep convolutional networks for large-scale image recognition. arXiv preprint arXiv:1409.1556.

He, K., Zhang, X., Ren, S. and Sun, J. (2016) Deep residual learning for image recognition, pp. 770-778.

Long, J., Shelhamer, E. and Darrell, T. (2015) Fully convolutional networks for semantic segmentation, pp. 3431-3440.

Badrinarayanan, V., Kendall, A. and Cipolla, R. (2017) Segnet: A deep convolutional encoder-decoder architecture for image segmentation. IEEE Transactions on Pattern Analysis and Machine Intelligence 39(12), 2481-2495.

Ronneberger, O., Fischer, P. and Brox, T. (2015) U-net: Convolutional networks for biomedical image segmentation, pp. 234-241, Springer.

Chen, L.-C., Zhu, Y., Papandreou, G., Schroff, F. and Adam, H. (2018) Encoder-decoder with atrous separable convolution for semantic image segmentation, pp. 801-818.

Zhao, X., Yuan, Y., Song, M., Ding, Y., Lin, F., Liang, D. and Zhang, D. (2019) Use of Unmanned Aerial Vehicle Imagery and Deep Learning UNet to Extract Rice Lodging. Sensors 19(18), 3859.

Dong, Z., Wang, G., Amankwah, S.O.Y., Wei, X., Hu, Y. and Feng, A. (2021) Monitoring the summer flooding in the Poyang Lake area of China in 2020 based on Sentinel-1 data and multiple convolutional neural networks. International Journal of Applied Earth Observation and Geoinformation 102, 102400.

Muhadi, N.A., Abdullah, A.F., Bejo, S.K., Mahadi, M.R. and Mijic, A. (2021) Deep Learning Semantic Segmentation for Water Level Estimation Using Surveillance Camera. Applied Sciences 11(20), 9691.

Hochreiter, S. and Schmidhuber, J. (1997) Long short-term memory. Neural computation 9(8), 1735-1780.

Oksuz, I., Cruz, G., Clough, J., Bustin, A., Fuin, N., Botnar, R.M., Prieto, C., King, A.P. and Schnabel, J.A. (2019) Magnetic Resonance Fingerprinting Using Recurrent Neural Networks, pp. 1537-1540.

Mellit, A., Pavan, A.M. and Lughi, V. (2021) Deep learning neural networks for short-term photovoltaic power forecasting. Renewable Energy 172, 276-288.

Gao, S., Huang, Y., Zhang, S., Han, J., Wang, G., Zhang, M. and Lin, Q. (2020) Short-term runoff prediction with GRU and LSTM networks without requiring time step optimization during sample generation. Journal of Hydrology 589, 125188.

Cho, M., Kim, C., Jung, K. and Jung, H. (2022) Water Level Prediction Model Applying a Long Short-Term Memory (LSTM)–Gated Recurrent Unit (GRU) Method for Flood Prediction. Water 14(14), 2221.

Schuster, M. and Paliwal, K.K. (1997) Bidirectional recurrent neural networks. IEEE

transactions on Signal Processing 45(11), 2673-2681.

Cho, K., Van Merriënboer, B., Gulcehre, C., Bahdanau, D., Bougares, F., Schwenk, H. and Bengio, Y. (2014) Learning phrase representations using RNN encoder-decoder for statistical machine translation. arXiv preprint arXiv:1406.1078.

Graves, A. and Schmidhuber, J. (2005) Framewise phoneme classification with bidirectional LSTM and other neural network architectures. Neural networks 18(5-6), 602-610.

McCuen, R.H., Knight, Z. and Cutter, A.G. (2006) Evaluation of the Nash-Sutcliffe Efficiency Index. Journal of Hydrologic Engineering 11(6), 597-602.

Gupta, H.V., Kling, H., Yilmaz, K.K. and Martinez, G.F. (2009) Decomposition of the mean squared error and NSE performance criteria: Implications for improving hydrological modelling. Journal of Hydrology 377(1), 80-91.

Knoben, W.J., Freer, J.E. and Woods, R.A. (2019) Inherent benchmark or not? Comparing Nash–Sutcliffe and Kling–Gupta efficiency scores. Hydrology and Earth System Sciences 23(10), 4323-4331.

Moriasi, D.N., Arnold, J.G., Van Liew, M.W., Bingner, R.L., Harmel, R.D. and Veith, T.L. (2007) Model evaluation guidelines for systematic quantification of accuracy in watershed simulations. Transactions of the ASABE 50(3), 885-900.

Zhou, Q.Q., Leng, G.Y. and Huang, M.Y. (2018) Impacts of future climate change on urban flood volumes in Hohhot in northern China: benefits of climate change mitigation and adaptations. Hydrology and Earth System Sciences 22(1), 305-316.

Zhou, Q., Ren, Y., Xu, M., Han, N. and Wang, H. (2016) Adaptation to urbanization impacts on drainage in the city of Hohhot, China. Water Science and Technology 73(1), 167-175.

MOHURD (2014) Technical Guidelines for Establishment of Intensity-Duration-Frequency Curve and Design Rainstorm Profile (In Chinese). Ministry of Housing and Urban-Rural Development of the People's Republic of China and China Meteorological Administration.

Berggren, K., Packman, J., Ashley, R. and Viklander, M. (2014) Climate changed rainfalls for urban drainage capacity assessment. Urban Water Journal 11(7), 543-556.

MIKE by DHI (2016) MIKE by DHI software, Release Note_MIKE URBAN, 2016, Horsolm, Denmark.

Kaspersen, P.S., Ravn, N.H., Arnbjerg-Nielsen, K., Madsen, H. and Drews, M. (2017) Comparison of the impacts of urban development and climate change on exposing European cities to pluvial flooding. Hydrology and Earth System Sciences 21(8), 4131-4147.

Zhou, Q., Mikkelsen, P.S., Halsnæs, K. and Arnbjerg-Nielsen, K. (2012) Framework for economic pluvial flood risk assessment considering climate change effects and adaptation benefits. Journal of Hydrology 414, 539-549.